\documentclass[11pt,a4paper]{article}
\usepackage[hyperref]{emnlp-ijcnlp-2019}
\usepackage{times}
\usepackage{latexsym}

\usepackage{url}

\usepackage{CJKutf8}
\usepackage{enumitem}
\usepackage{latexsym}
\usepackage{multirow}
\usepackage{amsmath,graphicx}
\usepackage{url}
\usepackage[export]{adjustbox}
\aclfinalcopy 

\definecolor{Alizarin}{rgb}{0.90, 0.1, 0.26}
\definecolor{Seeblue}{rgb}{0.28, 0.51, 0.95}

\title{Clickbait? Sensational Headline Generation with Auto-tuned Reinforcement Learning}
\author{Peng Xu, Chien-Sheng Wu, Andrea Madotto {\normalfont and} Pascale Fung  \\
  Center for Artificial Intelligence Research (CAiRE) \\
  Department of Electronic and Computer Engineering\\
  The Hong Kong University of Science and Technology, Clear Water Bay \\
  {\tt [pxuab,cwuak,amadotto]@connect.ust.hk, pascale@ece.ust.hk}
  \\}

\date{}

\begin{document}
\maketitle






\begin{abstract}
Sensational headlines are headlines that capture people's attention and generate reader interest. Conventional abstractive headline generation methods, unlike human writers, do not optimize for maximal reader attention. In this paper, we propose a model that generates sensational headlines without labeled data. We first train a sensationalism scorer by classifying online headlines with many comments (``clickbait'') against a baseline of headlines generated from a summarization model. The score from the sensationalism scorer is used as the reward for a reinforcement learner. However, maximizing the noisy sensationalism reward will generate unnatural phrases instead of sensational headlines. To effectively leverage this noisy reward, we propose a novel loss function, Auto-tuned Reinforcement Learning (ARL), to dynamically balance reinforcement learning (RL) with maximum likelihood estimation (MLE). Human evaluation shows that  60.8\% of samples generated by our model are sensational, which is significantly better than the Pointer-Gen baseline~\cite{see2017get} and other RL models.


\end{abstract}



\section{Introduction}




Headline generation is the process of creating a headline-style sentence given an input article. The research community has been regarding the task of headline generation as a summarization task \cite{shen2017recent}, ignoring the fundamental differences between headlines and summaries. While summaries aim to contain most of the important information from the articles, headlines do not necessarily need to. Instead, a good headline needs to capture people's attention and serve as an irresistible invitation for users to read through the article. For example, the headline ``\$2 Billion Worth of Free Media for Trump'', which gives only an intriguing hint, is considered better than the summarization style headline ``Measuring Trump’s Media Dominance'' \footnote{https://www.nytimes.com/2016/06/13/insider/which-headlines-attract-most-readers.html?module=inline}, as the former gets almost three times the readers as the latter. Generating headlines with many clicks is especially important in this digital age, because many of the revenues of journalism come from online advertisements and getting more user clicks means being more competitive in the market. However, most existing websites \footnote{http://www.contentrow.com/tools/link-bait-title-generator/} naively generate sensational headlines using only keywords or templates. Instead, this paper aims to learn a model that generates sensational headlines based on an input article without labeled data. 

To generate sensational headlines, there are two main challenges. Firstly, there is a lack of sensationalism scorer to measure how sensational a headline is. 
Some researchers have tried to manually label headlines as clickbait or non-clickbait \cite{chakraborty2016stop, potthast2018clickbait}. However, these human-annotated datasets are usually small and expensive to collect. To capture a large variety of sensationalization patterns, we need a cheap and easy way to collect a large number of sensational headlines. Thus, we propose a distant supervision strategy to collect a sensationalism dataset. We regard headlines receiving lots of comments as sensational samples and the headlines generated by a summarization model as non-sensational samples. Experimental results show that by distinguishing these two types of headlines, we can partially teach the model a sense of being sensational.

Secondly, after training a sensationalism scorer on our sensationalism dataset, a natural way to generate sensational headlines is to maximize the sensationalism score using reinforcement learning (RL). 
However, the following shows an example of a RL model maximizing the sensationalism score by generating a very unnatural sentence, while its sensationalism scorer gave a very high score of 0.99996: \begin{CJK*}{UTF8}{gbsn}
   \itshape 十个可穿戴产品的设计原则这消息消息可惜说明~ Ten design principles for wearable devices, this message message pity introduction. 
\end{CJK*}
This happens because the sensationalism scorer can make mistakes and RL can generate unnatural phrases which fools our sensationalism scorer. 
Thus, how to effectively leverage RL with noisy rewards remains an open problem. To deal with the noisy reward, we introduce Auto-tuned Reinforcement Learning (ARL). Our model automatically tunes the ratio between MLE and RL based on how sensational the training headline is. In this way, we effectively take advantage of RL with a noisy reward to generate headlines that are both sensational and fluent.


The major contributions of this paper are as follows: 
1) To the best of our knowledge, we propose the first-ever model that tackles the sensational headline generation task with reinforcement learning techniques. 
2) Without human-annotated data, we propose a distant supervision strategy to train a sensationalism scorer as a reward function.
3) We propose a novel loss function, Auto-tuned Reinforcement Learning, to give dynamic weights to balance between MLE and RL. Our code will be released .~\footnote{\url{https://github.com/HLTCHKUST/sensational_headline}} 








\section{Sensationalism Scorer}
To evaluate the sensationalism intensity score $\alpha_{\text{sen}}$ of a headline, we collect a sensationalism dataset and then train a sensationalism scorer. For the sensationalism dataset collection, we choose headlines with many comments from popular online websites as positive samples. For the negative samples, we propose to use the generated headlines from a sentence summarization model. Intuitively, the summarization model, which is trained to preserve the semantic meaning, will lose the sensationalization ability and thus the generated negative samples will be less sensational than the original one, similar to the obfuscation of style after back-translation \cite{prabhumoye2018style}. For example, an original headline like \begin{CJK*}{UTF8}{gbsn}``一趟挣10万？铁总增开申通、顺丰专列" (One trip to earn 100 thousand? China Railway opens new Shentong and Shunfeng special lines)\end{CJK*} will become \begin{CJK*}{UTF8}{gbsn}``中铁总将增开京广两列快递专列" (China Railway opens two special lines for express) \end{CJK*} from the baseline model, which loses the sensational phrases of \begin{CJK*}{UTF8}{gbsn}``一趟挣10万？" (One trip to earn 100 thousand?) \end{CJK*}. We then train the sensationalism scorer by classifying sensational and non-sensational headlines using a one-layer CNN with a binary cross entropy loss $L_{\text{sen}}$. Firstly, 1-D convolution is used to extract word features from the input embeddings of a headline. This is followed by a ReLU activation layer and a max-pooling layer along the time dimension. All features from different channels are concatenated together and projected to the sensationalism score by adding another fully connected layer with sigmoid activation. Binary cross entropy is used to compute the loss $L_{\text{sen}}$.


\subsection{Training Details and Dataset}
For the CNN model, we choose filter sizes of 1, 3, and 5 respectively. Adam is used to optimize $L_{sen}$ with a learning rate of 0.0001. We set the embedding size as 300 and initialize it from \newcite{qiu2018revisiting} trained on the Weibo corpus with word and character features. We fix the embeddings during training.  

For dataset collection, we utilize the headlines collected in \newcite{qin2018automatic, lin2019learning} from Tencent News, one of the most popular Chinese news websites, as the positive samples. We follow the same data split as the original paper. As some of the links are not available any more, we get 170,754 training samples and 4,511 validation samples. For the negative training samples collection,  we randomly select generated headlines from a pointer generator~\cite{see2017get} model trained on LCSTS dataset~\cite{hu2015lcsts} and create a balanced training corpus which includes 351,508 training samples and 9,022 validation samples. To evaluate our trained classifier, we construct a test set by randomly sampling 100 headlines from the test split of LCSTS dataset and the labels are obtained by 11 human annotators. Annotations show that 52\% headlines are labeled as positive and 48\% headlines as negative by majority voting (The detail on the annotation can be found in Section \ref{sec:eval_metrics}). 

\subsection{Results and Discussion}
Our classifier achieves 0.65 accuracy and 0.65 averaged F1 score on the test set while a random classifier would only achieve 0.50 accuracy and 0.50 averaged F1 score. This confirms that the predicted sensationalism score can partially capture the sensationalism of headlines. On the other hand, a more natural choice is to take headlines with few comments as negative examples. Thus, we train another baseline classifier on a crawled balanced sensationalism corpus of 84k headlines where the positive headlines have at least 28 comments and the negative headlines have less than 5 comments. However, the results on the test set show that the baseline classifier gets 60\% accuracy, which is worse than the proposed classifier (which achieves 65\%). The reason could be that the balanced sensationalism corpus are sampled from different distributions from the test set and it is hard for the trained model to generalize. Therefore, we choose the proposed one as our sensationalism scorer. 
Therefore, our next challenge is to show that how to leverage this noisy sensationalism reward to generate sensational headlines.

\begin{figure}[t]
    \centering
  \includegraphics[width=\linewidth,right]{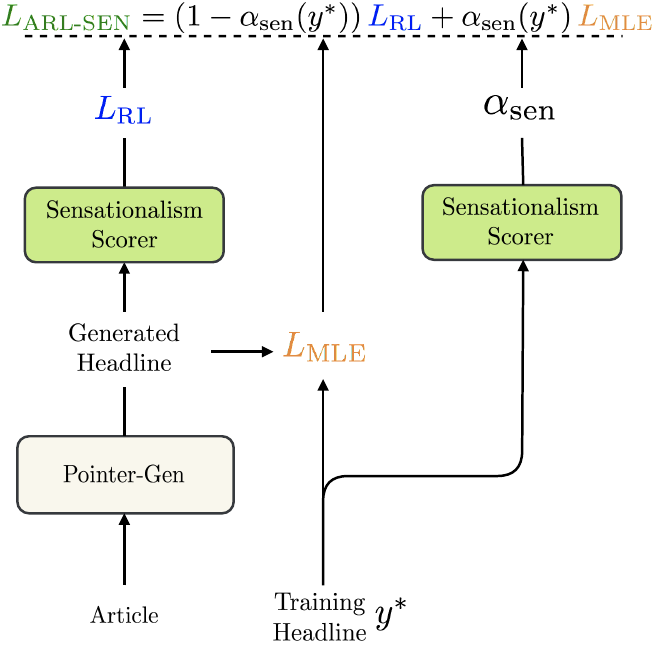}
  \caption{The loss function of Auto-tuned Reinforcement Learning is a weighted sum of $L_\text{RL}$ and $L_\text{MLE}$, where the weight is decided by our sensationalism scorer.}
  \label{fig:arl}
\end{figure}

\section{Sensational Headline Generation}
\label{sec:methodology}
Our sensational headline generation model takes an article as input and output a sensational headline. The model consists of a Pointer-Gen headline generator and is trained by ARL. The diagram of ARL can be found in Figure \ref{fig:arl}.

We denote the input article as $x=\{x_1,x_2,x_3,\cdots,x_M\}$, and the corresponding headline as $y^*=\{y_1^*,y_2^*,y_3^*,\cdots,y_T^*\}$, where $M$ is the number of tokens in an article and $T$ is the number of tokens in a headline.

\subsection{Pointer-Gen Headline Generator}
We choose Pointer Generator (Pointer-Gen)~\cite{see2017get}, a widely used summarization model, as our headline generator for its ability to copy words from the input article. It takes a news article as input and generates a headline. Firstly, the tokens of each article, $\{x_1,x_2,x_3,\cdots,x_M\}$, are fed into the encoder one-by-one and the encoder generates a sequence of hidden states $h_i$. For each decoding step $t$, the decoder receives the embedding for each token of a headline $y_t$ as input and updates its hidden states $s_t$. An attention mechanism following \newcite{luong2015effective} is used:
\begin{align}
    e_i^t &= v^T \text{tanh}(W_h h_i + W_s s_t + b_{attn}) \\
    a^t &= \text{softmax}(e^t) \\
    h_t^* &= \sum\nolimits_i a_i^t h_i
\end{align}
where $v$, $W_h$, $W_s$, and $b_{attn}$ are the trainable parameters and $h_t^*$ is the context vector. $s_t$ and $h_t^*$ are then combined to give a probability distribution over the vocabulary through two linear layers:
\begin{align}
    o_t & = V ([s_t, h_t^*]) + b \\
    P_{\text{voc}}(w) & = \text{softmax} ( V^{'} o_t + b^{'})
\end{align}
where $V$, $b$, $V^{'}$, and $b^{'}$ are trainable parameters. 
We use a pointer generator network to enable our model to copy rare/unknown words from the input article, giving the following final word probability:
\begin{align}
    p_{\text{gen}} & = \sigma (w_{h^*}^T h_t^* + w_s^T s_t + w_x^T x^t + b_{ptr}) \\
    P(w) & = p_{\text{gen}} P_{\text{voc}}(w) + (1 - p_{\text{gen}}) \sum_{i:x_i=w} a_i^t
\end{align}
where $x^t$ is the embedding of the input word of the decoder, $w_{h^*}^T$, $w_s^T$, $w_x^T$, and $b_{ptr}$ are trainable parameters, and $\sigma$ is the sigmoid function.

\subsection{Training Methods}
We first briefly introduce MLE and RL objective functions, and a naive way to mix these two by a hyper-parameter $\lambda$. Then we point out the challenge of training with noisy reward, and propose ARL to address this issue. 

\subsubsection{MLE and RL}
A headline generation model can be trained with MLE, RL or a combination of MLE and RL. MLE training is to minimize the negative log likelihood of the training headlines. We feed $y^*$ into the decoder word by word and maximize the likelihood of $y^*$. The loss function for MLE becomes
\begin{align} \label{eq:mle}
    L_{\text{MLE}} = - \frac{1}{T} \sum\nolimits_{t=1}^T \text{log} P(y_t^*) 
\end{align}

For RL training, we choose the REINFORCE algorithm \cite{williams1992simple}. In the training phase, after encoding an article, a headline $y^s = \{y_1^s, y_2^s, y_3^s, \cdots, y_T^s\}$ is obtained by sampling from $P(w)$ from our generator, and then a reward of sensationalism or ROUGE(RG) is calculated. 
\begin{align}
    R_{\text{rg}} &= \text{RG}(y^s, y^*) \\
    R_{\text{sen}} &= \alpha_{\text{sen}}(y^s)
\end{align}
We use the baseline reward $\hat{R_t}$ to reduce the variance of the reward, similar to \newcite{ranzato2015sequence}. To elaborate, a linear model is deployed to estimate the baseline reward $\hat{R_t}$ based on $t$-th  state $o_t$ for each timestep $t$. The parameters of the linear model are trained by minimizing the mean square loss between $R$ and $\hat{R_t}$:
\begin{align} \label{eq:baseline_reward}
\hat{R_t} &= W_r  o_t + b_r \\
L_b &= \frac{1}{T} \sum\nolimits_{t=1}^T | R - \hat{R_t} |^2
\end{align}
where $W_r$ and $b_r$ are trainable parameters. 
To maximize the expected reward, our loss function for RL becomes
\begin{align}  \label{eq:rl}
    L_{\text{RL}} = - \frac{1}{T} \sum\nolimits_{t=1}^T  (R - \hat{R_t}) \, \text{log} P(w_t) 
\end{align}

A naive way to mix these two objective functions using a hyper-parameter $\lambda$ has been successfully incorporated in the summarization task \cite{paulus2017deep}. 
It includes the MLE training as a language model to mitigate the readability and quality issues in RL. 
The mixed loss function is shown as follows:
\begin{equation}  \label{eq:rl-*}
    L_{\text{RL-*}} =  \lambda  L_{\text{RL}} +  (1 - \lambda)  L_{\text{MLE}}
\end{equation}
where $*$ is the reward type. Usually $\lambda$ is large, and \newcite{paulus2017deep} used 0.9984.

\subsubsection{Auto-tuned Reinforcement Learning}
Applying the naive mixed training method using sensationalism score as the reward is not obvious/trivial in our task. The main reason is that our sensationalism reward is notably more noisy and more fragile than the ROUGE-L reward or abstractive reward used in the summarization task \cite{paulus2017deep, kryscinski2018improving}. A higher ROUGE-L F1 reward in summarization indicates higher overlapping ratio between generation and true summary statistically, but our sensationalism reward is a learned score which is fragile to be fooled with unnatural samples. 

To effectively train the model with RL under noisy sensationalism reward, our idea is to balance RL with MLE. However, we argue that \textit{the weighted ratio between MLE and RL should be sample-dependent,} instead of being fixed for all training samples as in \newcite{paulus2017deep, kryscinski2018improving}. The reason is that, RL and MLE have inconsistent optimization objectives. When the training headline is non-sensational, MLE training will encourage our model to imitate the training headline (thus generating \textit{non-sensational} headlines), which counteracts the effects of RL training to generate \textit{sensational} headlines. 



The sensationalism score is, therefore, used to give dynamic weight to MLE and RL. 
Our ARL loss function becomes: 
\begin{align}  \label{eq:arl}
    L_{\text{ARL-SEN}} = (1 - \alpha_{\text{sen}}(y^*)) \, L_{\text{RL}} + \alpha_{\text{sen}}(y^*) \, L_{\text{MLE}}
\end{align}
If $\alpha_{\text{sen}}(y^*)$ is high, meaning the training headline is sensational, our loss function encourages our model to imitate the sample more using the MLE training. If $\alpha_{\text{sen}}(y^*)$ is low, our loss function replies on RL training to improve the sensationalism. Note that the weight $\alpha_{\text{sen}}(y^*)$ is different from our sensationalism reward $\alpha_{\text{sen}}(y^s)$ and we call the loss function Auto-tuned Reinforcement Learning, because the ratio between MLE and RL are well ``tuned'' towards different samples.  

\subsection{Dataset}
\label{sec:dataset}
We use LCSTS \cite{hu2015lcsts} as our dataset to train the summarization model. The dataset is collected from the Chinese microblogging website Sina Weibo. It contains over 2 million Chinese short texts with corresponding headlines given by the author of each text. The dataset is split into 2,400,591 samples for training, 10,666 samples for validation and 725 samples for testing. We tokenize each sentence with Jieba \footnote{https://github.com/fxsjy/jieba} and a vocabulary size of 50000 is saved. 

\subsection{Baselines and Our Models}
We experiment and compare with the following models.
\noindent\textbf{Pointer-Gen}  is the baseline model trained by optimizing $L_\text{MLE}$ in Equation \ref{eq:mle}.

\noindent\textbf{Pointer-Gen+Pos}  is the baseline model by training Pointer-Gen only on positive examples whose sensationalism score is larger than 0.5

\noindent\textbf{Pointer-Gen+Same-FT}  is the model which fine-tunes Pointer-Gen on the training samples whose sensationalism score is larger than 0.1

\noindent\textbf{Pointer-Gen+Pos-FT}  is the model which fine-tunes Pointer-Gen on the training samples whose sensationalism score is larger than 0.5

\noindent\textbf{Pointer-Gen+RL-ROUGE} is the baseline model trained by  optimizing $L_\text{RL-ROUGE}$ in Equation \ref{eq:rl-*}, with ROUGE-L \cite{lin2004rouge} as the reward.

\noindent\textbf{Pointer-Gen+RL-SEN} is the baseline model trained by optimizing $L_\text{RL-SEN}$ in Equation \ref{eq:rl-*}, with $\alpha_\text{sen}$ as the reward.

\noindent\textbf{Pointer-Gen+ARL-SEN} is our model trained by optimizing $L_\text{ARL-SEN}$ in Equation \ref{eq:arl}, with $\alpha_\text{sen}$ as the reward.

\noindent\textbf{Test set} is the headlines from the test set.

Note that we didn't compare to Pointer-Gen+ARL-ROUGE as it is actually Pointer-GEN. Recall that $\alpha_{\text{sen}}(y^*)$ in Equation \ref{eq:arl} measures how good (based on reward function) is $y^*$. Then the loss function for Pointer-Gen+ARL-ROUGE will be $$(1 - \text{RG}(y^*, y^*)) L_{\text{RL}} + \text{RG}(y^*,y^*) L_{\text{MLE}} =  L_{\text{MLE}}$$ We also tried text style transfer baseline \cite{shen2017style}, but the generated headlines were very poor (many unknown words and irrelevant). 

\begin{figure}[t]
  \centering
  \includegraphics[width=\linewidth]{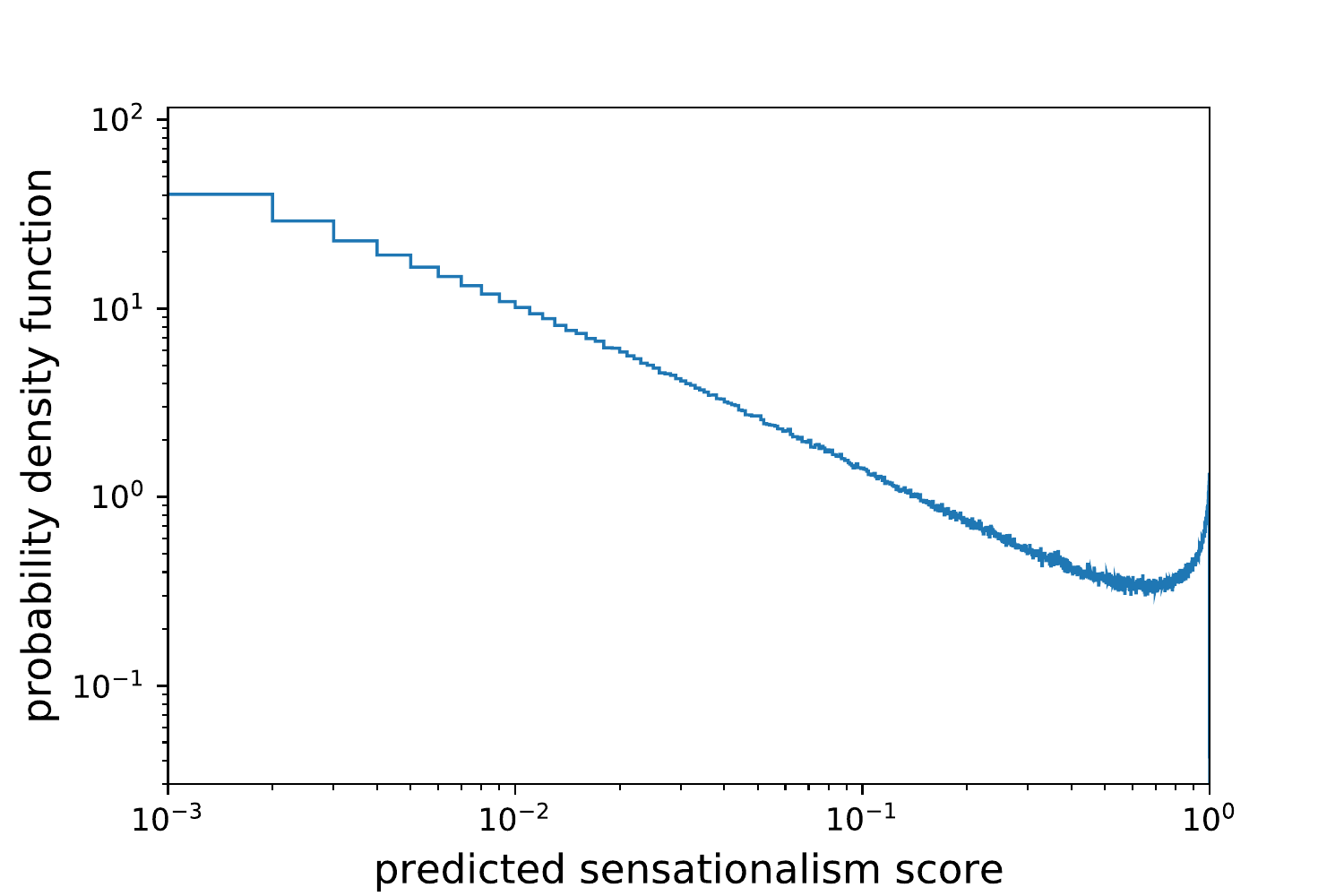}
  \caption{The probability density function (pdf) of predicted sensationalism score in log scale. Low sensationalism score has much higher probability density.}
  \label{fig:cdf_sensation}
\end{figure}

\subsection{Training Details}
\noindent\textbf{MLE training:} An Adam optimizer is used with the learning rate of 0.0001 to optimize $L_{\text{MLE}}$. The batch size is set as 128 and a one-layer, bi-directional Long Short-Term Memory (bi-LSTM) model with 512 hidden sizes and a 350 embedding size is utilized. Gradients with the l2 norm larger than 2.0 are clipped.  We stop training when the ROUGE-L f-score stops increasing. 

\noindent\textbf{Hybrid training:} An Adam optimizer with a learning rate of 0.0001 is used to optimize $L_{\text{RL-*}}$ (Equation \ref{eq:rl-*}) and $L_\text{{ARL-SEN}}$ (Equation \ref{eq:arl}).  When training Pointer-Gen+RL-ROUGE, the best $\lambda$ is chosen based on the ROUGE-L score on the validation set. In our experiment, $\lambda$ is set as 0.95. An Adam optimizer with a learning rate of 0.001 is used to optimize $L_b$. When training Pointer-Gen+ARL-SEN, we don't use the full LCSTS dataset, but only headlines with a sensationalism score larger than 0.1 as we observe that Pointer-Gen+ARL-SEN will generate a few unnatural phrases when using full dataset. We believe the reason is the high ratio of RL during training. Figure \ref{fig:cdf_sensation} shows that the probability density near 0 is very high, meaning that in each batch, many of the samples will have a very low sensationalism score. On expectation, each sample will receive 0.239 MLE training and 0.761 RL training. This leads to RL dominanting the loss. 
Thus, we propose to filter samples with a minimum sensationalism score with 0.1 and it works very well. For Pointer-Gen+RL-SEN, we also set the minimum sensationalism score as 0.1, and $\lambda$ is set as 0.5 to remove unnatural phrases, making a fair comparison to Pointer-Gen+ARL-SEN. 

We stop training Pointer-Gen+Same-FT, Pointer-Gen+Pos-FT, Pointer-Gen+RL-SEN and Pointer-Gen+ARL-SEN, when $\alpha_\text{sen}$ stops increasing on the validation set.  Beam-search with a beam size of 5 is adopted for decoding in all models. 


\begin{table}[t]
\centering
\resizebox{\linewidth}{!}{
\begin{tabular}{l|l|l|l}
\hline
                              & \bf{RG-1}  & \bf{RG-2}  & \bf{RG-L}  \\ \hline
character-based preprocessing & \multicolumn{3}{l}{}           \\ \hline
RNN context~\cite{hu2015lcsts}        & 29.90    & 17.40    & 27.20    \\ 
COPYNET~\cite{gu2016incorporating}  & 34.40    & 21.60    & 31.30    \\ \hline
word-based preprocessing      & \multicolumn{3}{l}{}           \\ \hline
RNN context~\cite{hu2015lcsts}           & 26.80    & 16.10    & 24.10    \\
COPYNET~\cite{gu2016incorporating}    & 35.00    & 22.30    & 32.00    \\
Pointer-Gen                   & 34.51    & 22.21    & 31.68    \\
Pointer-Gen+Pos               & 28.51    & 16.53    & 25.56 \\
Pointer-Gen+Same-FT                   & 34.60    & 22.00    & 31.48    \\
Pointer-Gen+Pos-FT            & 30.92    & 18.76    & 28.02    \\
Pointer-Gen+RL-ROUGE          & \bf{36.26}    & \bf{23.48}    & \bf{33.21}    \\
Pointer-Gen+RL-SEN          & 35.06   & 22.37   & 31.91    \\
Pointer-Gen+ARL-SEN               & 34.28    & 21.34    & 30.80   \\ \hline
\end{tabular}}
\caption{\label{table: rouge comparison} Our implementation achieves similar performance to the RNN context and COPYNET. Pointer-Gen+ARL-SEN achieves good summarization performance even though it is optimized for the sensational reward. It shows its ability to summarize.}
\end{table}

\subsection{Evaluation Metrics}
\label{sec:eval_metrics}
We briefly describe the evaluation metrics below. \\
\noindent\textbf{ROUGE:} ROUGE is a commonly used evaluation metric for summarization. It measures the N-gram overlap between generated and training headlines. We use it to evaluate the \textit{relevance} of generated headlines. The widely used pyrouge~\footnote{https://github.com/bheinzerling/pyrouge} toolkit is used to calculate ROUGE-1 (RG-1), ROUGE-2 (RG-2), and ROUGE-L (RG-L).

\noindent\textbf{Human evaluation:} We randomly sample 50 articles from the test set and send the generated headlines from all models and corresponding headlines in the test set to human annotators. We evaluate the sensationalism and fluency of the headlines by setting up two independent human annotation tasks. We ask 10 annotators to label each headline for each task. For the sensationalism annotation, each annotator is asked one question, ``Is the headline sensational?'', and he/she has to choose either `yes' or `no'. The annotators were not told which system the headline is from. The process of distributing samples and recruiting annotators is managed by Crowdflower.\footnote{https://www.figure-eight.com/} After annotation, we define the sensationalism score as the proportion of annotations on all generated headlines from one model labeled as `yes'. For the fluency annotation, we repeat the same procedure as for the sensationalism annotation, except that  we ask each annotator the question ``Is the headline fluent?'' We define the fluency score as the proportion of annotations on all headlines from one specific model labeled as `yes'. We put human annotation instructions in the supplemental material.

\begin{CJK*}{UTF8}{gbsn}
\begin{table}[t]
\begin{center}
\resizebox{\linewidth}{!}{
\begin{tabular}{|l|}
\hline  \textbf{Article}: 
昨天的央视315晚会上，尼康D600相机被曝拍摄\\
时会出现黑色斑点，反复修理也无果，而尼康竟把责任\\
推给了雾霾，拒绝换机或退机。对此上海工商部门昨晚\\
连夜梳理了近年对尼康的投诉案，今天上午前往位于黄\\
浦区的尼康公司进一步调查。新民网\\
At the CCTV 315 party yesterday, the Nikon D600 camera is \\
reported to have black spots when taking photos. Repeated \\
repairs gave no results, and Nikon actually attributes the\\
damage to the smog, refusing exchange or return. The \\
Shanghai Industrial and Commercial Department collected \\
the recent complaints against Nikon last night, and went to \\
Nikon Company in Huangpu District for further investigation\\
this morning. Xinmin Net\\
\hline \textbf{Pointer-Gen: }  尼康D600相机被曝拍摄时会出现黑色斑点 \\
The Nikon D600 camera is reported to have black spots when \\taking photos \\
\hline \textbf{Pointer-Gen+Same-FT: }  尼康D600被曝拍出雾霾尼康已\\
介入调查 The Nikon D600 camera is reported to have smog \\
Nikon started investigation \\
\hline \textbf{Pointer-Gen+Pos-FT: }  尼康投诉央视315晚会尼康投诉被\\
曝 Nikon complains CCTV 315 night, Nikon is reported to \\
be complained\\
\hline \textbf{Pointer-Gen+RL-ROUGE}: 尼康D600相机被曝拍摄时会\\
出现黑色斑点 \quad  The Nikon D600 camera is reported to have \\
black spots when taking photos \\
\hline \textbf{Pointer-Gen+RL-SEN}: 尼康D600被曝拍出奇葩事尼康把\\
责任推给雾霾 \quad The Nikon D600 camera is reported to have \\
something strange and attributes the damage to the smog.\\
\hline \textbf{Pointer-Gen+ARL-SEN (Ours)}: \textbf{\textcolor{Seeblue}{摊上大事了！}} 尼康D600\\
被指拍出``黑点” \textbf{\textcolor{Seeblue}{In Serious Trouble!}} The Nikon D600 \\
camera is reported to have black spots when taking photos \\
\hline \textbf{Test set headline}: 尼康称黑点怪雾霾上海工商今上门追查\\
Nikon attributes black spots to the smog, and Shanghai Indus-\\
trial and Commercial Department start investigation today. \\
\hline
\end{tabular}}
\end{center}
\caption{\label{table: sensational_example} Generated Chinese headlines from different models. Our model~(Pointer-Gen+ARL-SEN) sensationalized the headline with the phrase ``In Serious Trouble!".}  
\end{table}
\end{CJK*}

\begin{table}[t]
\centering
\resizebox{\linewidth}{!}{
\begin{tabular}{c|c|c}
\hline
    Model & sensationalism & fluency \\ \hline
    Pointer-Gen & 42.6\%* & 80\% \\
    Pointer-Gen-Pos & 40.2\%* & 59\%* \\
    Pointer-Gen+Same-FT & 47.6\%*  & 75.6\% \\
    Pointer-Gen+Pos-FT & 47.8\%* &  71.4\% \\
    Pointer-Gen+RL-ROUGE   & 45.2\%*  & 80\% \\
    Pointer-Gen+RL-SEN     & 51.8\%*        & 79.4\% \\
    Pointer-Gen+ARL-SEN & \bf{60.8\%}  & 79.4\% \\ \hline
    Test set  & 57.8\% & \bf{80.6\%}\\ \hline

\end{tabular}}
\caption{\label{table: human evaluation} Comparison of sensationalism score and fluency score between different models. Pointer-Gen+ARL-SEN achieves the best performance among all models in sensationalism score. * indicates Pointer-Gen+ARL-SEN is statistically significantly better than the corresponding model.}
\end{table}



\section{Results}

We first compare all four models, Pointer-Gen, Pointer-Gen-RL+ROUGE, Pointer-Gen-RL-SEN, and Pointer-Gen-ARL-SEN, to existing models with ROUGE in Table \ref{table: rouge comparison} to establish that our model produces relevant headlines and we leave the sensationalism for human evaluation. Note that we only compare our models to commonly used strong summarization baselines, to validate that our implementation achieves comparable performance to existing work. In our implementation, Pointer-Gen achieves a 34.51 RG-1 score, 22.21 RG-2 score, and 31.68 RG-L score, which is similar to the results of \newcite{gu2016incorporating}. Pointer-Gen+ARL-SEN, although optimized for the sensationalism reward, achieves similar performance to our Pointer-Gen baseline, which means that Pointer-Gen+ARL-SEN still keeps its summarization ability. An example of headlines generated from different models in Table \ref{table: sensational_example} shows that Pointer-Gen and  Pointer-Gen+RL-ROUGE learns to summarize the main point of the article: ``The Nikon D600 camera is reported to have black spots when taking photos''.  Pointer-Gen+RL-SEN makes the headline more sensational by blaming Nikon for attributing the damage to the smog.  Pointer-Gen+ARL-SEN generates the most sensational headline by exaggerating the result  ``Getting a serious trouble!'' to maximize user's attention. 

\begin{figure}[t]
  \centering
  \includegraphics[width=\linewidth]{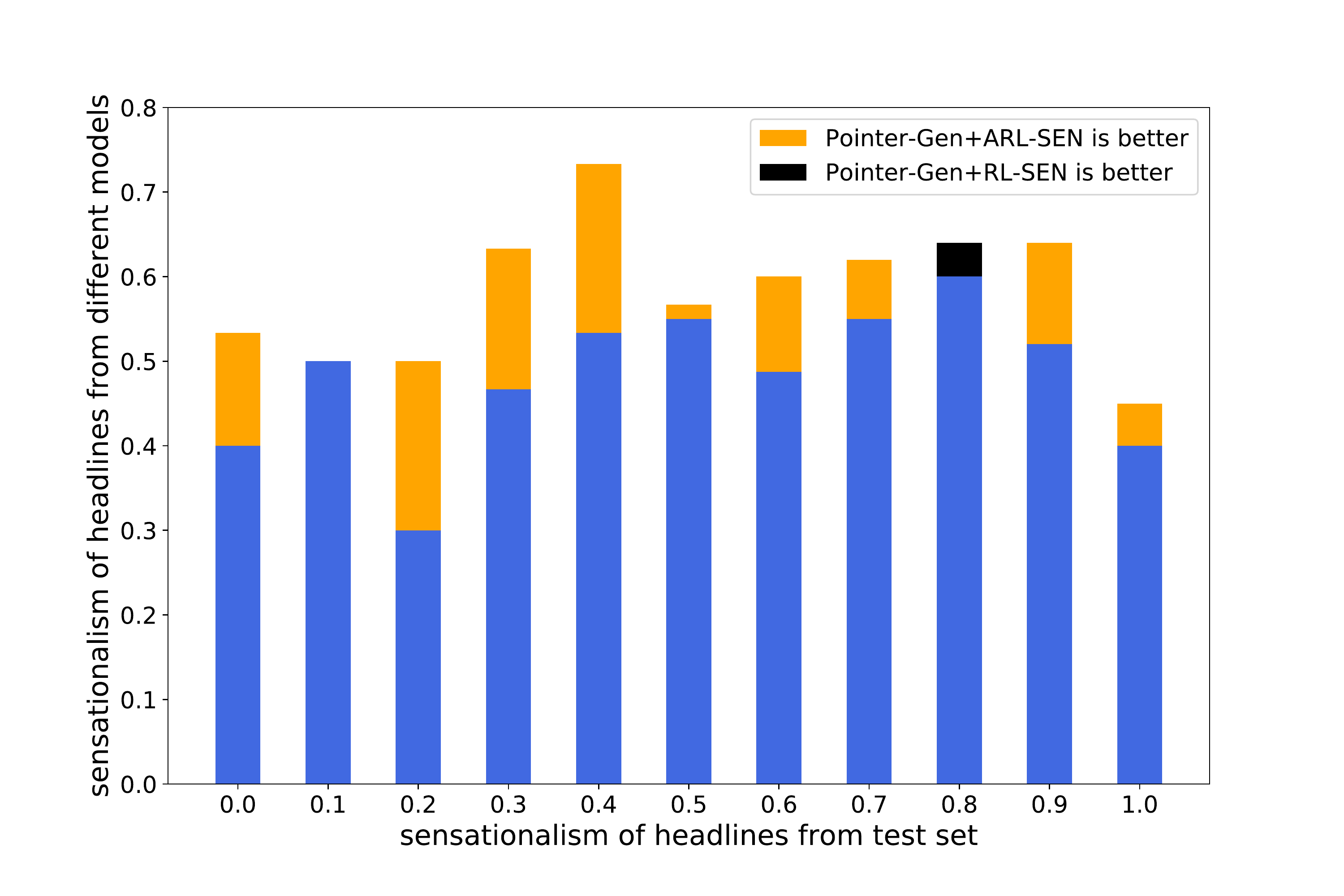}
  \caption{Comparison of sensationalism score between Pointer-Gen+ARL-SEN and Pointer-Gen+RL-SEN for different test set headlines. The blue bars denote the smaller scores between the two models. Pointer-Gen+ARL-SEN achieves better performance on most cases. Greater improvement is achieved when the test set headline is non-sensational.}
  \label{fig:arl_rl_comparison}
\end{figure}

We then compare different models using the sensationalism score in Table \ref{table: human evaluation}.  The Pointer-Gen baseline model achieves a 42.6\% sensationalism score, which is the minimum that a typical summarization model achieves. By filtering out low-sensational headlines, Pointer-Gen+Same-FT and Pointer-Gen+Pos-FT achieves higher sensationalism scores, which implies the effectiveness of our sensationalism scorer. Our Pointer-Gen+ARL-SEN model achieves the best performance of 60.8\%. This is an absolute improvement of 18.2\% over the Pointer-Gen baseline. The Chi-square test on the results confirms that Pointer-Gen+ARL-SEN is statistically significantly more sensational than all the other baseline models, with the largest p-value less than 0.01. Also, we find that the test set headlines achieves 57.8\% sensationalism score, much larger than Pointer-Gen baseline, which also supports our intuition that generated headlines will be less sensational than the original one. On the other hand, we found that Pointer-Gen+Pos is much worse than other baselines. The reason is that training on sensational samples alone discards around 80\% of the whole training set that is also helpful for maintaining relevance and a good language model. It shows the necessity of using RL.

\begin{CJK*}{UTF8}{gbsn}
\begin{table*}[t]
\begin{center}
\resizebox{\linewidth}{!}{
\begin{tabular}{|c|l|}
\hline
\multirow{8}{*}{Question}               & \textbf{Pointer-Gen}:    眺望：印度经济复苏的陷阱 \\  & View: the trap of economic recovery in India                                                                                           \\
                                                                     
                                                                      &\textbf{ Pointer-Gen+ARL-SEN}:   印度\textbf{\textcolor{Seeblue}{或将}}成为下一个中国 \textbf{\textcolor{Seeblue}{?}} \\  & \textbf{\textcolor{Seeblue}{Will}} India \textbf{\textcolor{Seeblue}{become}} the next China\textbf{\textcolor{Seeblue}{?}}    \\           \cline{2-2}                                          
                                                & \textbf{Pointer-Gen}:  今天理论导报的主要内容有……                                             \\
                                                                      & The main content of today’s theoretical report is                                                                          \\ 
                                                                      & \textbf{Pointer-Gen+ARL-SEN}:   \textbf{\textcolor{Seeblue}{如何}}做一名焦裕禄式的县委书记\textbf{\textcolor{Seeblue}{?}}                                                  \\
                                                                      & \textbf{\textcolor{Seeblue}{How to}} be a county party secretary like JIAO Yulu\textbf{\textcolor{Seeblue}{?}}                                                        \\ \hline
\multirow{7}{*}{Creating Curiosity Gap} & \textbf{Pointer-Gen}: 10种方法帮你避免的10个小窍门                                                                                           \\
                                                                      & 10 tricks to help you avoid                                                 \\  
                                                                      & \textbf{Pointer-Gen+ARL-SEN}: 10个让你\textbf{\textcolor{Seeblue}{意想不到}}的领导力法则                                                              \\
                                                                      & 10 laws of leadership that you \textbf{\textcolor{Seeblue}{cannot think of}} \\ \cline{2-2}
                                                                    &  \textbf{Pointer-Gen:} 王林的暴富之路 \quad WANG Lin's path to sudden wealth                                 \\  
                                                                      & \textbf{Pointer-Gen+ARL-SEN}: “气功大师”王林的暴富之路：凭借\textbf{\textcolor{Seeblue}{5个生财之道}}                                                              \\
                                                                      & ``The Qigong Master" WANG Lin's path to sudden wealth: leveraging \textbf{\textcolor{Seeblue}{5 ways to make money}} \\ \hline
\multirow{8}{*}{Highlighting Numbers}     & \textbf{Pointer-Gen}: 北京市集成电路促进基金就位                                                                                        \\
                                                                      & The Integrated Circuit Promotion Fund in Beijing is ready                                                             \\ 
                                                                      & \textbf{Pointer-Gen+ARL-SEN}: \textbf{\textcolor{Seeblue}{500亿巨资}}驰援国家大基金或已就位                                                                  \\
                                                                      &  \textbf{\textcolor{Seeblue}{A huge capital sum of 50 billion}} is ready to support the national big fund              \\ \cline{2-2} 
                                          & \textbf{Pointer-Gen}:   
                                          陈光标冰桶挑战陈光标承认造假\\
                                                                      & CHEN Guangbiao admits that he cheated in the freezing water challenge                                                              \\  
                                                                      & \textbf{Pointer-Gen+ARL-SEN}: 陈光标回应``造假''：有人超越我就捐款 \textbf{\textcolor{Seeblue}{100万}}                                        \\
                                                                      & CHEN Guangbiao responded to ``cheating'': if someone can do better, I will donate \textbf{\textcolor{Seeblue}{a million RMB}}               \\ \hline                   
                                        
\multirow{8}{*}{Emotional Words}               & \textbf{Pointer-Gen}: 俞永福：搜狗360还没停  \\
& YU Yongfu: Sougou and 360 have not stopped                                                      \\ 
                                                                      & \textbf{Pointer-Gen+ARL-SEN}: 俞永福微博\textbf{\textcolor{Seeblue}{辱骂}}UC：\textbf{\textcolor{Seeblue}{太恶心了}}                                                                 \\
                                                                      & YU Youfu abused UC in microblog: it is too  \textbf{\textcolor{Seeblue}{disgusting}}                           \\ \cline{2-2}
                                               
                                                                      & \textbf{Pointer-Gen}: 保定楼市````一天涨3000"" \\
                                                                      & ``3000 increase in one day" for Baoding housing market                             \\ 
                                                                       & \textbf{Pointer-Gen+ARL-SEN}: 保定楼市“一天涨3000 \textbf{\textcolor{Seeblue}{简直疯了}}”                                                           \\
                                                                       & ``3000 increase in one day" for Baoding housing market, \textbf{\textcolor{Seeblue}{this is crazy}}                          \\ \cline{2-2}\hline
 \multirow{7}{*}{Empathizing}               & \textbf{Pointer-Gen}: 女性购物的五大特征 \quad Five characteristics of ladies shopping
                                                 \\ 
                                                                       & \textbf{Pointer-Gen+ARL-SEN}: 女性购物的5个忠告：\textbf{\textcolor{Seeblue}{你}}中枪了吗？                                                                \\
                                                                       &  5 warnings for ladies shopping: Have  \textbf{\textcolor{Seeblue}{you}} been targeted?                      \\ \cline{2-2}
                                                               & \textbf{Pointer-Gen}: 智能手表将开始拥有智能手机功能                                                                            \\
                                                               &  Smart watches will start to have smartphone features                                                   \\
                                                                      & \textbf{Pointer-Gen+ARL-SEN}: 关于智能手表，\textbf{\textcolor{Seeblue}{你}}应该知道的事！                                         \\  
                                                                      & What \textbf{\textcolor{Seeblue}{you}} should know about smart watches!                                                                        \\
                                                                       \hline

\end{tabular}}
\end{center}
\caption{\label{table: sensational examples} Different sensationalization strategies Pointer-Gen+ARL-SEN learns. } 
\end{table*}
\end{CJK*}


In addition, 
both Pointer-Gen+RL-SEN and Pointer-Gen+ARL-SEN, which use the sensationalism score as the reward, obtain statistically better results than Pointer-Gen+RL-ROUGE and Pointer-Gen, with a p-value less than 0.05 by a Chi-square test. This result shows the effectiveness of RL to generate more sensational headlines. The reason is that even though our noisy classifier could also learn to classify domains, the generator during RL training is not allowed to increase the reward by shifting domains but encouraged to generate more sensational headlines, due to the consistency constraint on the domains of the headline and the article. Furthermore, Poiner-Gen+ARL-SEN gets better performance than Pointer-Gen+RL-SEN, which confirms the superiority of the ARL loss function. We also visualize in Figure \ref{fig:arl_rl_comparison} a comparison between Pointer-Gen+ARL-SEN and Pointer-Gen+RL-SEN according to how sensational the test set headlines are. The blue bars denote the smaller scores between the two models. For example, if the blue bar is 0.6, it means that the worse model between Pointer-Gen+RL-SEN and Pointer-Gen+ARL-SEN achieves 0.6. And the color of orange/black further indicates the better model and its score. We find that Pointer-Gen+ARL-SEN outperforms Pointer-Gen+RL-SEN for most cases. The improvement is higher when the test set headlines are not sensational (the sensationalism score is less than 0.5), which may be attributed to the higher ratio of RL training on non-sensational headlines. 



Apart from the sensationalism evaluation, we measure the fluency of the headlines generated from different models. Fluency scores in Table \ref{table: human evaluation} show that Pointer-Gen+RL-SEN and Pointer-Gen+ARL-SEN achieve comparable fluency performance to Pointer-Gen and Pointer-Gen+RL-ROUGE. Test set headlines achieve the best performance among all models, but the difference is not statistically significant. Also, we observe that fine-tuning on sensational headlines will hurt the performance, both in sensationalism and fluency.

After manually checking the outputs, we observe that our model is able to generate sensational headlines using diverse sensationalization strategies. These strategies include, but are not limited to, creating a curiosity gap, asking questions, highlighting numbers, being emotional and emphasizing the user. Examples can be found in Table \ref{table: sensational examples}.

\section{Related Work}
\label{sec:related_work}

Our work is related to summarization tasks. An encoder-decoder model was first applied to two sentence-level abstractive summarization tasks on the DUC-2004 and  Gigaword datasets \cite{rush2015neural}. This model was later extended by selective encoding \cite{zhou2017selective}, a coarse to fine approach \cite{tan2017neural}, minimum risk training \cite{shen2017recent}, and topic-aware models \cite{wang2018reinforced}. As long summaries were recognized as important, the CNN/Daily Mail dataset was used in \newcite{nallapati2016abstractive}. Graph-based attention~\cite{tan2017abstractive}, pointer-generator with coverage loss \cite{see2017get} are further developed to improve the generated summaries. 
\newcite{celikyilmaz2018deep} proposed deep communicating agents for representing a long document for abstractive summarization. 
In addition, many papers \cite{nallapati2017summarunner, zhou2018neural,zhang2018neural} use extractive methods to directly select sentences from articles.
However, none of these work considered the sensationalism of generated outputs.

RL is also gaining popularity as it can directly optimize non-differentiable metrics \cite{pasunuru2018multi, venkatraman2015improving, xu2019novel}. \newcite{paulus2017deep}  proposed an intra-decoder model and combined RL and MLE to deal with summaries with bad qualities. RL has also been explored with generative adversarial networks (GANs) \cite{yu2017seqgan}. \newcite{liu2017generative} applied GANs on summarization task and achieved better performance. \newcite{niu2018polite} tackles the problem of polite generation with politeness reward. Our work is different in that we propose a novel function to  balance RL and MLE.

Our task is also related to text style transfer. 
Implicit methods~\cite{shen2017style, fu2018style,prabhumoye2018style} transfer the styles by separating sentence representations into content and style, for example using back-translation\cite{prabhumoye2018style}. However, these methods cannot guarantee the content consistency between the original sentence and transferred output~\cite{xu2018unpaired}. Explicit methods~\cite{zhang2018learning, xu2018unpaired} transfer the style by directly identifying style related keywords and modifying them. However, sensationalism is not always restricted to keywords, but the full sentence. By leveraging small human labeled English dataset, clickbait detection has been well investigated \cite{chakraborty2016stop,shu2018deep,potthast2018clickbait}. However, these human labeled dataset are not available for other languages, such as Chinese. 

Modeling sensationalism is also related to modeling emotion. Emotion has been well investigated in both word level\cite{Tang2016,xu2018emo2vec} and sentence level\cite{Felbo2017, winata2019caire_hkust, winata2018attention, park2018plusemo2vec, lee2019team}. It has also been considered an important factor in engaging interactive systems\cite{lin2019caire, winata2017nora,zhou2018design}. Although we observe that sensational headlines contain emotion, it is still not clear which emotion and how emotions will influence the sensationalism. 

\section{Conclusion and Future Work}
In this paper, we propose a model that generates sensational headlines without labeled data using Reinforcement Learning. Firstly, we propose a distant supervision strategy to train the sensationalism scorer. As a result, we achieve 65\% accuracy between the predicted sensationalism score and human evaluation.  To effectively leverage this noisy sensationalism score as the reward for RL, we propose a novel loss function, ARL, to automatically balance RL with MLE. Human evaluation confirms the effectiveness of both our sensationalism scorer and ARL to generate more sensational headlines. Future work can be improving the sensationalism scorer and investigating the applications of dynamic balancing methods between RL and MLE in textGAN\cite{yu2017seqgan}. Our work also raises the ethical questions about generating sensational headlines, which can be further explored.

\section*{Acknowledgments}
Thanks to ITS/319/16FP of Innovation Technology Commission, HKUST 16248016 of Hong Kong Research Grants Council for funding. In addition, we thank Zhaojiang Lin for helpful discussion and Yan Xu, Zihan Liu for the data collection.



\bibliography{emnlp-ijcnlp-2019}
\bibliographystyle{acl_natbib}

\clearpage

\end{document}